\documentclass{article} 
\usepackage{iclr2023_conference,times}

\usepackage{amsmath,amsfonts,bm}

\def\eqref#1{equation~\ref{#1}}

\def\1{\bm{1}}

\def\rveps{{\mathbf{\epsilon}}}

\def\rvc{{\mathbf{c}}}
\def\rvd{{\mathbf{d}}}

\def\rvo{{\mathbf{o}}}
\def\rvp{{\mathbf{p}}}

\def\rvr{{\mathbf{r}}}

\def\rvv{{\mathbf{v}}}

\def\rvx{{\mathbf{x}}}

\def\rvz{{\mathbf{z}}}

\def\rmI{{\mathbf{I}}}

\def\rmV{{\mathbf{V}}}

\def\vzero{{\bm{0}}}

\DeclareMathAlphabet{\mathsfit}{\encodingdefault}{\sfdefault}{m}{sl}
\SetMathAlphabet{\mathsfit}{bold}{\encodingdefault}{\sfdefault}{bx}{n}

\def\gN{{\mathcal{N}}}

\def\sR{{\mathbb{R}}}

\newcommand{\E}{\mathbb{E}}
\newcommand{\Ls}{\mathcal{L}}

\newcommand\paren[1]{\left(#1\right)}
\newcommand\abs[1]{\left\lvert#1\right\rvert}
\newcommand\norm[1]{\left\lVert#1\right\rVert}

\usepackage{hyperref}
\usepackage{url}

\title{Learning a Diffusion Prior for NeRFs}

\author{Guandao Yang$^1$, Abhijit Kundu$^2$, Leonidas J. Guibas$^2$, Jonathan T. Barron$^2$, Ben Poole$^2$\\
$^1$Cornell University, $^2$Google Research
}

\usepackage{booktabs}
\usepackage{xcolor}
\usepackage{graphicx}
\usepackage[export]{adjustbox}
\usepackage{lipsum}
\usepackage{algorithm}
\usepackage{algpseudocode}
\usepackage{tabularx}
\usepackage{wrapfig}

\newcommand{\ignore}[1]{}

\iclrfinalcopy 
\begin{document}

\maketitle

\begin{abstract}
Neural Radiance Fields (NeRFs) have emerged as a powerful neural 3D representation for objects and scenes derived from 2D data. 
Generating NeRFs, however, remains difficult in many scenarios.
For instance, training a NeRF with only a small number of views as supervision remains challenging since it is an under-constrained problem.
In such settings, it calls for some inductive prior to filter out bad local minima.
One way to introduce such inductive priors is to learn a generative model for NeRFs modeling a certain class of scenes.
In this paper, we propose to use a diffusion model to generate NeRFs encoded on a regularized grid.
We show that our model can sample realistic NeRFs, while at the same time allowing conditional generations, given a certain observation as guidance.
\end{abstract}

\section{Introduction}\label{sec:intro}

Neural Radiance Fields (NeRFs)~\cite{NeRF,ReLUField,Instant-NGP,DVGO,Plenoxels} have emerged as a popular neural scene representation thanks to their ability to represent 3D scenes with  high fidelity. 
While NeRFs excel for 3D reconstruction tasks where many views are available, novel view synthesis from NeRFs with few views remains a challenging problem. At training time, the NeRF is only rendered from observed views, and there is no prior enforcing how the NeRF should behave in areas that are unobserved.

A generative model of NeRFs of certain categories can potentially address such ambiguities.
A generative model approximates the data distribution by producing samples that resemble those sampled from the data distribution or by estimating how likely a test data can appear according to the training data distribution.
For example, a NeRF generative model of cars can create NeRFs whose rendered views are indistinguishable from real-world car scenes.
Such a generative model contains information to rule out badly reconstructed car scenes as these scenes are unlikely to be created by the model.
To achieve this, a generative model for NeRFs should be able to capture the data distribution well (i.e., generate realistic NeRFs or provide an estimate of the likelihood of a proposed NeRF).
Moreover, it also needs to be capable of performing conditional generation since downstream applications often need to generate NeRFs that can explain different forms of input data, such as posed images or sparse point clouds.


In this paper, we take a step toward this goal by adapting the recently popular diffusion models~\cite{sohl2015deep,DDPM,ScoreSDE} to generate NeRFs that are encoded with spatially supported features (e.g., ReLU-fields~\cite{ReLUField}).
Diffusion models have shown the ability to generate state-of-the-art quality in a variety of signal generation tasks, including images and videos~\cite{diffusionbeatsgan,ho2022video}.
Such models synthesize a sample by iteratively refining the generated samples according to a trained network.
This iterative procedure also provides the flexibility to adapt an unconditional generative model and make it aware of various conditioning signals~\cite{ScoreSDE,graikos2022diffusion,nichol2021improved,ho2022video}.
These two properties of the diffusion model suggest that diffusion model can be a good solution to both generate high-quality NeRFs and also support test-time optimization with different conditional inputs.
Specifically, we propose a mechanism to create a dataset of NeRFs represented in ReLU-fields that's suitable to be learned by diffusion models.
We provide empirical results demonstrating that the diffusion model trained with a dataset generated by our mechanism achieves good generative performance.
We also show that NeRF diffusion model can be used as a prior to improve the single-view 3D reconstruction task.

\section{Related Works}

Our methods are connected to three bodies of prior works: learning NeRFs from few views, generative models for NeRF, and diffusion models. 
We discuss the most relevant related works in the following sections.

\textbf{Generative Models for NeRFs:}
Currently, the most successful NeRF generative models usually apply GAN techniques~\cite{GAN} to generate NeRFs~\cite{EG3D,GRAF,EpiGRAF}.
While these GAN-based methods are capable of sampling high-quality rendered images unconditionally, it is non-trivial to leverage this generative prior in conditional generation tasks.
This is partially due to the difficulty of inverting a GAN decoder or of using GAN to evaluate the likelihood.
Alternatively, prior works have also attempted using VAEs~\cite{VAE} to generate NeRFs~\cite{NeRFVAE,ProbNeRF}.
Such variational inference provides an estimate of likelihood and thus provision to perform conditional generation with various data.
These methods usually fail to generate NeRFs with high-fidelity rendered qualities, partially caused by the fact that their latent code dimension is low.
It remains challenging to design generative NeRF models that supports not only high-quality unconditional samples but also facilitate conditional generation.

\textbf{Diffusion Models and Score-based Generative Models:} 
Diffusion and score-based generative models have been shown to be capable of producing high-quality results in multiple tasks, including image generation~\cite{DDPM,ScoreSDE,diffusionbeatsgan}, image-to-image translation~\cite{Palatte}, image super resolution~\cite{S3}, text-to-images~\cite{LDM}, and videos~\cite{ho2022video}.
Of particular relevance to our work are works that apply diffusion models to generate representations of 3D shapes and studies to condition the generation procedure on certain guidance.
Prior works have tried to apply diffusion and score-based generative models to generate 3D point clouds~\cite{ShapeGF,PointVoxDiff}.
Recently, GAUDI~\cite{GAUDI} and Functa~\cite{functa} applied diffusion model to learn the distribution of the global latent code that got decoded into a NeRF.
These works have to share the decoder between the generative model and the rendering decoder, making the rendering procedure difficult to run fast.
In our paper, we propose to apply diffusion to spatial features, which allows fast rendering.
In addition, our work also leverages guidance studied in diffusion model literature~\cite{ho2022video,plug-and-play,Dreamfusion} to make our unconditional generative model useful as a prior for different downstream applications. 
Concurrent works have explored similar ideas~\cite{zhang20233dshape2vecset,muller2022diffrf}.

\textbf{Conditional NeRF Generation:}
Per our discussion in Sec~\ref{sec:intro}, it's essential to encode prior in the algorithm to succeed in optimizing NeRF conditioned on certain signals.
One of the most studied conditional NeRF generation task is to reconstruct NeRF from few images.
The common way to impose prior for this task is by adding heuristic regularization terms to the test-time optimization objective~\cite{RegNeRF}.
LearnInit~\cite{LearnInit} and PlenopticNF~\cite{PNF} used meta-learning to learn an initialization for the test-time optimization.
Such prior doesn't guarantee the optimization to stay within the meaningful data manifold.
Another line of work tries to directly learn the mapping from image to NeRF~\cite{PixelNeRF,CodeNeRF}.
The abovementioned methods develop prior to the specific conditional signal.
Our proposed model can be used not only in the few-view reconstruction but also in applications with different conditioning signals, such as point clouds.





\section{Preliminary}\label{sec:background}

\textbf{NeRF}:
A Neural Radiance Field represents a scene using a function $F_\theta$ that takes spatial coordinate $\rvx \in \sR^3$ and a ray direction $\rvd \in \sR^2$ as input and outputs the tuple contains the volume density $\sigma \in \sR$ and emitted color in RGB $\rvc \in \sR^3$: $F_\theta(\rvx, \rvd) = (\rvc, \sigma)$.
The common way to obtain a NeRF from a collection of posed images is to optimize the loss between input and the volume rendered images.
The expected color from a ray $\rvr(t, \rvd) = \rvo + t \rvd$ can be computed using
$
    C_f(\rvd) =\int T(t)\sigma(\rvd(t))\rvc(\rvr(t,\rvd), \rvd)dt \label{eq:vol-render},
$,
where $T(t) = \exp\paren{-\int \sigma(\rvr(s))ds}$ is the accumulated transmittance along the ray.
Let $C(\rvd)$ denote the ground truth color along the ray from the training data.
We will compare the expected rendered color with the ground truth and optimize parameter $\theta$ with the following loss:
\begin{align}
    \mathcal{L}_{NeRF} = \sum_{\rvd} \norm{C(\rvd) - C_f(\rvd)}_2^2 \label{eq:NeRF-color}.
\end{align}

NeRFs are usually represented using MLP neural networks, which require a very long time to converge. 
Recent works have shown that representing NeRFs with spatially supported features can lead to much faster convergence~\cite{ReLUField,DVGO,Instant-NGP,Plenoxels}.
These methods commonly assume that the function $F_\theta$ takes the form of
$
    F_\theta(\rmV)(\rvx, \rvd) = f_\theta(\operatorname{interp}(\rmV, \rvx), \rvx, \rvd),
$
where $\rmV$ is a grid of features, $\operatorname{interp}$ usually refers to linear interpolation, and $f_\theta$ is a small neural network or deterministic function that takes the interpolated feature and outputs the tuple $(\rvc, \sigma)$.
In this paper, we settle on using ReLU-field~\cite{ReLUField} as NeRF backbone, a design decision which we will discuss in later section~\ref{sec:first-stage}.

\textbf{Diffusion Models:}
Diffusion model first defines a way to gradually perturb the data $\rvx$ using Gaussian noise: $q(\mathbf{z}_t|\rvx) =\gN(\alpha_t\rvx, \sigma_t^2 \rmI)$, where $\alpha_t$ and $\sigma_t^2$ are a positive-valued function of $t$.
Then $q(\mathbf{z}_t)$ defines a diffusion process as $t$ goes to $\infty$. 
The generative model is defined by inverting the diffusion process.
Assume that the signal-to-noise ratio $\alpha_t^2/\sigma_t^2$ is small enough at $t=1$, so  $q(\mathbf{z}_1|\mathbf{x}) \approx \gN(\vzero, \rmI)$.
We first discretize the time interval $[0, 1]$ into $T$ pieces.
Then we can define a hierarchical generative model for data $\rvx$ using
$
    p_\theta(\rvx) = \int_\rvz p(\rvz_1)p_\theta(\rvx|\rvz_0)\prod_{i=1}^T p_\theta(\rvz_{s(i)} | \rvz_{t(i)}),
$
where $s(i) = (i-1)/T$, $t(i) = i/T$, $\theta$ are the parameters of the neural network used to approximate distributions such as $p_\theta(\rvz_s|\rvz_t)$ and the likelihood $p_\theta(\rvx | \rvz_0)$.
If we parameterize $p_\theta(\rvz_s|\rvz_t) = q(\rvz_s|\rvz_t, \rvx=\hat{\rvx}_\theta(\rvz_t, t))$ with neural network $\rvx_\theta$, then we can train this generative model by optimizing the ELBO:
$
    D_{KL}(q(\rvz_1|\rvx)||p(\rvz_1)) + \E_{q(\rvz_0|\rvx)}\left[-\log p(\rvx|\rvz_0)\right] +\Ls_T,
$
where the diffusion loss $\Ls_T$ can be computed as:
\begin{align}
\Ls_T = \frac{T}{2}\E_{\rveps\sim\gN(\vzero, \rmI),t\sim U(0, T]} \left[w(t)\norm{\rvx-\hat{\rvx}_\theta(\rvx, t)}\right],
\end{align}
where $w(t)$ is the weight defined as $\frac{\alpha_{t+1}^2}{\sigma_{t+1}^2} - \frac{\alpha_t^2}{\sigma_t^2}$.
The common way to sample from the diffusion model is ancestral sampling:
\begin{align}
    \rvz_1 \sim \gN\paren{\vzero, \rmI}, \rvz_{s(i)} \sim p_\theta\paren{\rvz_{s(i)} | \rvz_{t(i)}}, \forall i \in [1, T]
    \rvx \sim p_\theta(\rvx|\rvz_0).
\end{align}
One nice property of this sampling algorithm is that it allows the incorporation of different forms of guidance to perform conditional sampling, which we will discuss in detail in Sec~\ref{sec:test-time-opt}.


\section{Methods}\label{sec:method}

From Section~\ref{sec:intro}, our goal is to build a diffusion generative prior for NeRFs that's capable of generating NeRFs and being used as a prior for different test-time optimization algorithms.
To train a diffusion prior for NeRFs, however, is non-trivial.
There can be multiple different NeRFs that represent the same scene, which creates an issue for the diffusion model.
To successfully generate a NeRF with a diffusion model, the network should be able to take a noisy NeRF representation and predict a less noisy version.
If there exist multiple plausible denoised results for one given signal, this can make the denoiser harder to train.

In this section, we will propose a method to tackle these problems.
We follow~\cite{LDM,PointFlow,ShapeGF} to break down the generation procedure into two stages: first stage will learn a latent representation of the scene, and the second stage will learn a generative model in that latent representation.
In our first stage, we propose to train regularized ReLU-field, a representation of NeRF that's suitable to learn a diffusion model.We will show that regularized ReLU fields are easy to obtain and they can be denoised using standard U-Net architecture since it's represented in a grid (Sec~\ref{sec:first-stage}).
In our second stage, we will train a DDPM model on the regularized ReLU-fields (Sec~\ref{sec:second-stage}). 
Finally, we propose a way to use reconstruction guidance~\cite{ho2022video} to perform test-time optimization for a variety of 3D tasks, including single-view NeRF reconstruction (Sec~\ref{sec:test-time-opt}).

\subsection{Regularized ReLU-F}\label{sec:first-stage}
\begin{figure*}[t]
  \centering
  \begin{tabular}{cc|cc}
      \multicolumn{2}{c}{Not regularized} & \multicolumn{2}{c}{Regularized} \\
      \includegraphics[width=0.1\linewidth,height=0.1\linewidth]{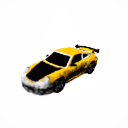} &
      \includegraphics[width=0.1\linewidth,height=0.1\linewidth]{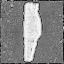}
      \includegraphics[width=0.1\linewidth,height=0.1\linewidth]{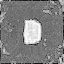}
      \includegraphics[width=0.1\linewidth,height=0.1\linewidth]{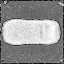} &
      \includegraphics[width=0.1\linewidth,height=0.1\linewidth]{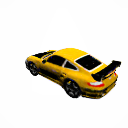} &
      \includegraphics[width=0.1\linewidth,height=0.1\linewidth]{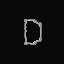}
      \includegraphics[width=0.1\linewidth,height=0.1\linewidth]{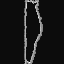}
      \includegraphics[width=0.1\linewidth,height=0.1\linewidth]{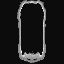} \\
      \includegraphics[width=0.1\linewidth,height=0.1\linewidth]{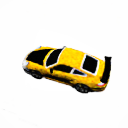} &
      \includegraphics[width=0.1\linewidth,height=0.1\linewidth]{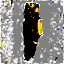}
      \includegraphics[width=0.1\linewidth,height=0.1\linewidth]{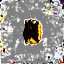}
      \includegraphics[width=0.1\linewidth,height=0.1\linewidth]{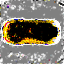} &
      \includegraphics[width=0.1\linewidth,height=0.1\linewidth]{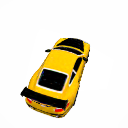} &
      \includegraphics[width=0.1\linewidth,height=0.1\linewidth]{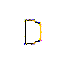}
      \includegraphics[width=0.1\linewidth,height=0.1\linewidth]{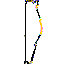}
      \includegraphics[width=0.1\linewidth,height=0.1\linewidth]{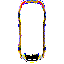}
  \end{tabular}
\vspace{-1em}
   \caption{Comparing ReLU-fields with regularized ReLU-fields. The left side shows ReLU-fields trained without regularization and the right shows our method.}
   \label{fig:first-stage}
\end{figure*}

To learn a diffusion generative prior for NeRF, it's essential to find the right representation of NeRFs suitable to be learned by diffusion model.
A good representation should have the following properties:
a) \textit{easy-to-obtain}. Training a diffusion model requires a lots of data, so we need our representation to be easy to obtain
b) \textit{structured}. In order for diffusion model to be able to denoise it progressively, we need the model to have structure, and
c) \textit{expressive}. Since we want to be able to represent signals in high fedelity.
Using the trainable weights of an MLP as the representation doesn't work since it's not structured enough to be denoised.
Representing NeRFs with a latent vector or grid used to modulate the MLP can be difficult to train since we need to syncrhonize gradients for the shared MLP.
With these considerations, we choose to represent NeRFs with ReLu fields because it's easy-to-obtain (i.e. takes about 2 minutes to obtain a scene in 1 GPU), expressive, and can be made structured.
Our first stage will try to obtain NeRFs in the form of ReLu fields.

We assume we have a dataset of $N$ scenes, each of which has $M$ posed images: $\{(\rvp_{ij}, I_{ij})\}_{j=1}^M\}_{i=1}^N$, where $\rvp_{ij}$ is the $j^{th}$ camera posed and pixel images for the $i^{th}$ scene.
For each scene $i$, we want to train a ReLU field using the posed images $\{(\rvp_{ij}, I_{ij})\}_{j=1}^N$.
A ReLU-field $F_\rmV:\sR^{3}\times \sR^2 \mapsto \sR \times \sR^3$, which takes a tuple of 3D coordinate and direction and output a tuple of density and RGB value.
$F_\rmV$ is parameterized by a grid of size $\rmV \in \sR^{R\times R \times R \times 4}$. 
Given this, we have $F_\rmV(\rvx, \rvd) = f(\operatorname{interp}(\rmV, \rvx), \rvx, \rvd)$, where $\operatorname{interp}$ is trilinear interpolation.
The function $f$ is parameter-less that takes the $4$-D interpolated feature to predict the final density and RGB value. 
The density is defined as $f(\rvv, \rvx, \rvd)_{den} = \exp(\alpha\rvv[0] + \beta)$ and the RGB color is obtained by $f(\rvv, \rvx, \rvd)_{rgb} = \operatorname{Sigmoid}(\rvv[1:4])$.
\footnote{While we ignore the effect of view-direction for simplicity, users can add it in by using spherical-harmonics encoding~\cite{Plenoxels}}.

The parameter $\rmV$ is optimized by loss objective showed in Equation~\ref{eq:NeRF-color}.
But naively optimizing this objective without any regularization will lead to an ill-posed problem as there are many different ways to instantiate the density and the color field to satisfied the photometric loss.
This is due to the fact that there are many voxels in the field that are either 1) \textit{free-space} (i.e. having very low-density value), and/or 2) \textit{occluded} (i.e., there is no ray that can create substantial alpha composition value for the color at that location).
Letting these invisible voxels to be set into arbitrary value does not affect the final rendered image, but it makes the data set more complex and less easy to denoise.
As shown in Figure~\ref{fig:first-stage}.
To alleviate this issue, we regularize these fields with the following two losses:

\textbf{Density sparsity regularization:} We encourage the field to have sparse density, so that it only put high density in the places where the color does matters. Let $d_{min}$ denotes a hyper-parameter of pre-activation minimal density such that $\exp(d_{min}\alpha + \beta) \approx 0$, the density sparsity regularization can be written as following:
\begin{align}
    \Ls_{d}(\rmV) = \norm{\rmV[..., 0] - d_{min}}_1,
\end{align}

\textbf{Color constancy regularization:} We encourage the color of the field to be as white as possible by putting a huber-loss between the voxel color to a white color:
\begin{align}
    \Ls_{c}(\rmV) = \begin{cases}
        \frac{1}{2}\norm{\rmV[..., 1:4] - c}_2 &\text{if } \abs{\rmV[..., 1:4]} < \delta \\
        \delta\paren{\abs{\rmV[..., 1:4] - c} - \frac{1}{2}\delta}\ & \text{otherwise}
    \end{cases},
\end{align}
where $c$ is the default color we want the albedo volume to have.
Currently we set it to color white.

Finally, our regularized ReLU-Fields are training with combining three losses:
$
    \Ls_d(\rmV) + \Ls_c(\rmV) + \sum_{j=1}^M \Ls_{NeRF}(\rmV, \rvp_{ij}, I_{ij}).
$
Figure~\ref{fig:first-stage} shows that with these regularization, we are able to obtain fields with more structured without sacrificing the quality and convergence speed. 

\subsection{Training a Diffusion Prior in NeRF}\label{sec:second-stage}
\begin{wrapfigure}{R}{0.4\textwidth}
    \vspace{-2.2em}
    \begin{minipage}{0.4\textwidth}
      \begin{algorithm}[H]
        \caption{Diffusion-Guided Test-time Optimization}\label{alg:opt}
        \begin{algorithmic}
        \Require $n \geq 0$
        \State $\rmV \sim \gN(\vzero, I)$
        \For{$i=1\dots T$}
        \State $\rmV_{t(i)} \gets \rmV$
        \State $\rmV \gets p_\theta(\rmV_{s(i)} | \rmV_{t(i)})$
        \For{$j=1\dots K$}
        \State $\rmV \gets \rmV + \alpha \nabla_{\rmV} \log{p_\theta(y|\rmV)}$
        \EndFor
        \EndFor
        \end{algorithmic}
        \end{algorithm}
    \end{minipage}
\end{wrapfigure}
The previous section has given us a set of NeRFs, parameterized by $\{\rmV_i\}_{i=0}^N$, where $f_{\rmV_i}$ is the radiance field for scene $i$.
We would like to learn a diffusion model $p_\theta(\rmV)$ to model the distribution of $q(\rmV)$, which we assumed $\rmV_i$ are sampled from.
We update the U-Net architecture of~\citet{ScoreSDE} by replacing the 2D convolution with 3D convolution.
3D convolution significantly increase the activation memory, so we downsample the $32^3$ fields three times to resolution of $4^3$.
We also apply self-attentions for 3D features grids in resolutions of $8$ and $4$.
We use $6$ residual blocks of width $128$ for each resolution.
We use Adam optimizer with learning rate $0.0001$ for $1M$ iterations with batch size $128$.
We apply gradient clipping of $500$ and warm up the learning rate with $5000$ iterations to avoid instability in training.


\subsection{Test-time Inference}\label{sec:test-time-opt}
Now that we've obtained a generative prior in the form of a diffusion model, so we are able to sample Relu-F volume $\rmV_0$ with ancestral sampling:
\begin{align}
    \rmV_1 \sim \gN(\vzero, I), \rmV_{s(i)} \sim p_\theta(\rmV_{s(i)} | \rmV_{t(i)}), \forall i \in [1, T].
\end{align}
To build a generative prior that's also capable of doing conditional sampling, we need to provide an algorithm to sample a ReLU-F volume that satisfies certain conditioning observation.
Specifically, given a conditional observation $y$ and a function that can produce the observation $p(y|\rmV)$, we would like to sample $\rmV$ from $p_\theta(\rmV|y)$.
This can be achieved by updating the same ancestral sampling with the conditional probability:
\begin{align}
    \rmV_1 \sim \gN(\vzero, I), \rmV_{s(i)} \sim p_\theta(\rmV_{s(i)} | \rmV_{t(i)}, y), \forall i \in [1, T].
\end{align}
The conditional distribution $p_\theta(\rmV_{s(i)} | \rmV_{t(i)}, y)$ can be estimated using Bayes' Rule:
\begin{align}
    p_\theta(\rmV_{s(i)} | \rmV_{t(i)}, y) = \frac{
    p_\theta(y | \rmV_{s(i)})p_\theta(\rmV_{s(i)} | \rmV_{t(i)})
    }{p_\theta(y | \rmV_{t(i)})} \propto p_\theta(y | \rmV_{s(i)})p_\theta(\rmV_{s(i)} | \rmV_{t(i)})\footnote{}.
\end{align}
This shows that as long as we are able to estimate guidance probability $p_\theta(y | \rmV_t)$, we are able to sample ReLU-F conditioned on the observation $y$.
There are many different ways to compute $p_\theta(y |\rmV_t)$~\cite{ho2022video,ScoreSDE}, many of which assume 

In the application of single-view reconstruction, the observation $y$ will be an image from a given pose $\rvp$,
One way to define $p_\theta(y|\rmV_t)$ is to directly use the learning objective $\Ls_{NeRF}$.
An alternative way to define $p_\theta(y|\rmV_t)$ is to use the denoised version of the volume $\hat{\rmV}$. 
In our test-time optimization algorithm, we will use both guidance.
Please see Algorithm~\ref{alg:opt} for more detail.

\section{Experiments}\label{sec:results}

In this section, we will show some preliminary results of our methods in unconditional generation and a conditional generation task - single-view reconstruction.
We mainly conduct experiments on the car category of the ShapeNet dataset~\cite{ShapeNet} and follow Scene Representation Network~\cite{SRN} to preprocess the data.
For each training scene in the dataset, we randomly sample $250$ different camera poses at the sphere and render $128\times 128$ resolution RGB images.
We obtain a regularized ReLU-field for each shape.
Obtaining one regularized ReLU-field takes 2 minutes with 8 V100 GPUs.

\subsection{Unconditional Generation}
\begin{figure*}[t]
  \centering
    \begin{tabular}{cc}
             \rotatebox{90}{\ \ No reg}  & 
        \adjincludegraphics[width=0.11\textwidth,height=0.11\textwidth,trim={0 0 {.5\width} {.5\height}},clip]{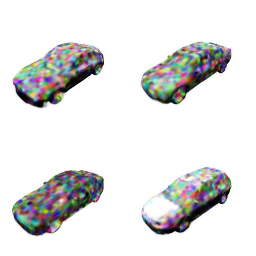}
        \adjincludegraphics[width=0.11\textwidth,height=0.11\textwidth,trim={{.5\width} 0 0 {.5\height}},clip]{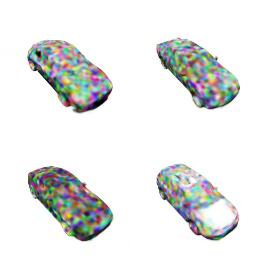}
        \adjincludegraphics[width=0.11\textwidth,height=0.11\textwidth,trim={0 {.5\height} {.5\width} 0 },clip]{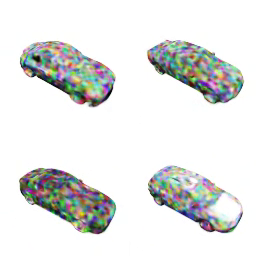}
        \adjincludegraphics[width=0.11\textwidth,height=0.11\textwidth,trim={{.5\width} {.5\height} 0 0},clip]{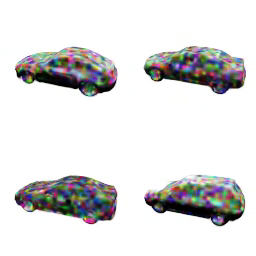}

        \adjincludegraphics[width=0.11\textwidth,height=0.11\textwidth,trim={0 0 {.5\width} {.5\height}},clip]{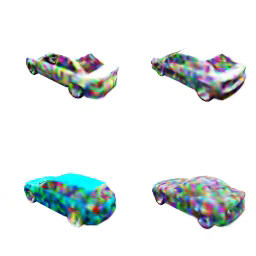}
        \adjincludegraphics[width=0.11\textwidth,height=0.11\textwidth,trim={{.5\width} 0 0 {.5\height}},clip]{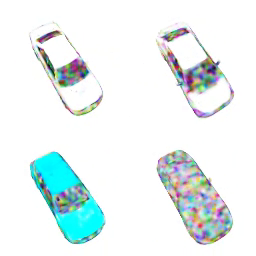}
        \adjincludegraphics[width=0.11\textwidth,height=0.11\textwidth,trim={0 {.5\height} {.5\width} 0 },clip]{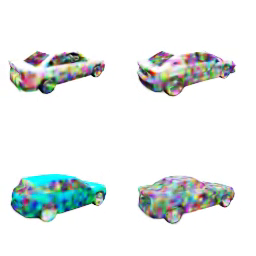}
        \adjincludegraphics[width=0.11\textwidth,height=0.11\textwidth,trim={{.5\width} {.5\height} 0 0},clip]{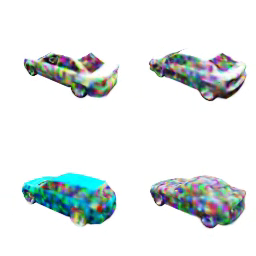}
        \\
        \rotatebox{90}{\ \ \ Ours} &
        \includegraphics[width=0.11\textwidth,height=0.11\textwidth]{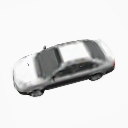}
        \includegraphics[width=0.11\textwidth,height=0.11\textwidth]{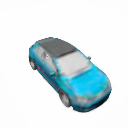}
        \includegraphics[width=0.11\textwidth,height=0.11\textwidth]{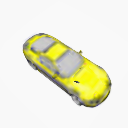}
        \includegraphics[width=0.11\textwidth,height=0.11\textwidth]{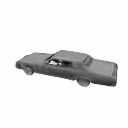}
        \includegraphics[width=0.11\textwidth,height=0.11\textwidth]{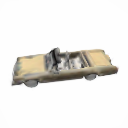}
        \includegraphics[width=0.11\textwidth,height=0.11\textwidth]{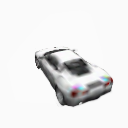}
        \includegraphics[width=0.11\textwidth,height=0.11\textwidth]{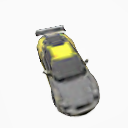}
        \includegraphics[width=0.11\textwidth,height=0.11\textwidth]{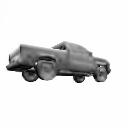} 
    \end{tabular}
\vspace{-1em}
\caption{Unconditional generation results. The top row shows samples of diffusion model trained with ReLU-fields without regularization. The bottom row shows samples of diffusion model trained with regularized ReLU-fields. Our model is able to produce more diverse shapes and correct colors.}
\label{fig:generation}
\end{figure*}

We first show that our model can capture the distribution of the NeRF dataset by examining the results of unconditional generation.
We compare our model to the baseline, which train the diffusion model on non-regularized ReLU-fields.
Figure~\ref{fig:generation} shows the generation results of our model.
Without regularizing the ReLU-fields, the diffusion model cannot capture the data distirbution. 
As shown in the first row, the baseline produces cars with averaged shapes and wonky colors.
On the other hand, our model is able to create cars with various shapes and correct color pattern.

\subsection{Single-view Reconstruction}
\begin{figure*}
\centering
\begin{tabular}{ccc}
&
Input &
Output 
\\
\rotatebox{90}{\ \ Ours} &
\includegraphics[width=0.1\linewidth,height=0.1\linewidth]{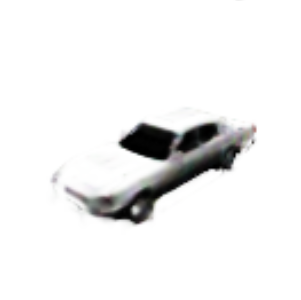} &
\adjincludegraphics[width=0.1\linewidth,height=0.1\linewidth,trim={0 0 {.875\width} 0},clip]{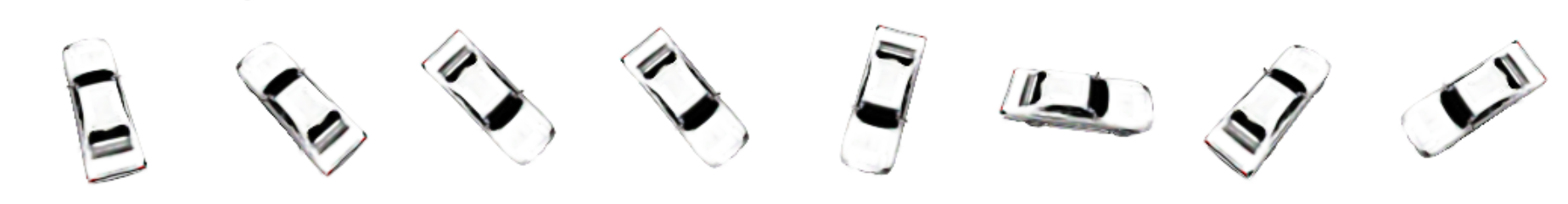}
\adjincludegraphics[width=0.1\linewidth,height=0.1\linewidth,trim={{.125\width} 0 {.75\width} 0},clip]{figures/few_view_recon/test-ours.png}
\adjincludegraphics[width=0.1\linewidth,height=0.1\linewidth,trim={{.25\width} 0 {.625\width} 0},clip]{figures/few_view_recon/test-ours.png}
\adjincludegraphics[width=0.1\linewidth,height=0.1\linewidth,trim={{.375\width} 0 {.5\width} 0},clip]{figures/few_view_recon/test-ours.png}
\adjincludegraphics[width=0.1\linewidth,height=0.1\linewidth,trim={{.5\width} 0 {.375\width} 0},clip]{figures/few_view_recon/test-ours.png}
\adjincludegraphics[width=0.1\linewidth,height=0.1\linewidth,trim={{.625\width} 0 {.25\width} 0},clip]{figures/few_view_recon/test-ours.png}
\adjincludegraphics[width=0.1\linewidth,height=0.1\linewidth,trim={{.75\width} 0 {.125\width} 0},clip]{figures/few_view_recon/test-ours.png}
\adjincludegraphics[width=0.1\linewidth,height=0.1\linewidth,trim={{.875\width} 0 {0\width} 0},clip]{figures/few_view_recon/test-ours.png}
\\
\rotatebox{90}{\quad\ Real} &
\includegraphics[width=0.1\linewidth,height=0.1\linewidth]{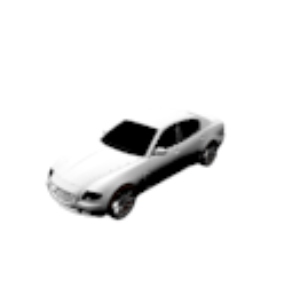} &
\adjincludegraphics[width=0.1\linewidth,height=0.1\linewidth,trim={0 0 {.875\width} 0},clip]{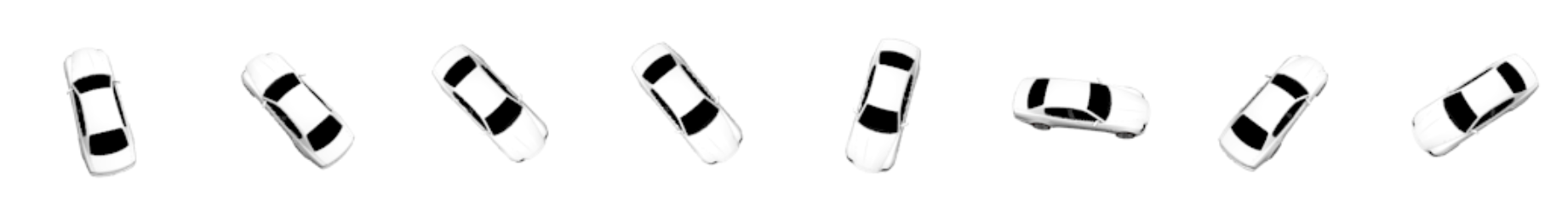}
\adjincludegraphics[width=0.1\linewidth,height=0.1\linewidth,trim={{.125\width} 0 {.75\width} 0},clip]{figures/few_view_recon/test-gtr.png}
\adjincludegraphics[width=0.1\linewidth,height=0.1\linewidth,trim={{.25\width} 0 {.625\width} 0},clip]{figures/few_view_recon/test-gtr.png}
\adjincludegraphics[width=0.1\linewidth,height=0.1\linewidth,trim={{.375\width} 0 {.5\width} 0},clip]{figures/few_view_recon/test-gtr.png}
\adjincludegraphics[width=0.1\linewidth,height=0.1\linewidth,trim={{.5\width} 0 {.375\width} 0},clip]{figures/few_view_recon/test-gtr.png}
\adjincludegraphics[width=0.1\linewidth,height=0.1\linewidth,trim={{.625\width} 0 {.25\width} 0},clip]{figures/few_view_recon/test-gtr.png}
\adjincludegraphics[width=0.1\linewidth,height=0.1\linewidth,trim={{.75\width} 0 {.125\width} 0},clip]{figures/few_view_recon/test-gtr.png}
\adjincludegraphics[width=0.1\linewidth,height=0.1\linewidth,trim={{.875\width} 0 {0\width} 0},clip]{figures/few_view_recon/test-gtr.png} \\
\hline
\rotatebox{90}{\ \ Ours} &
\includegraphics[width=0.1\linewidth,height=0.1\linewidth]{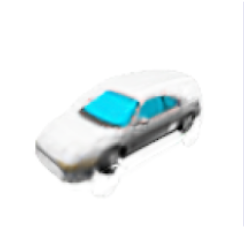} &
\adjincludegraphics[width=0.1\linewidth,height=0.1\linewidth,trim={0 0 {.875\width} 0},clip]{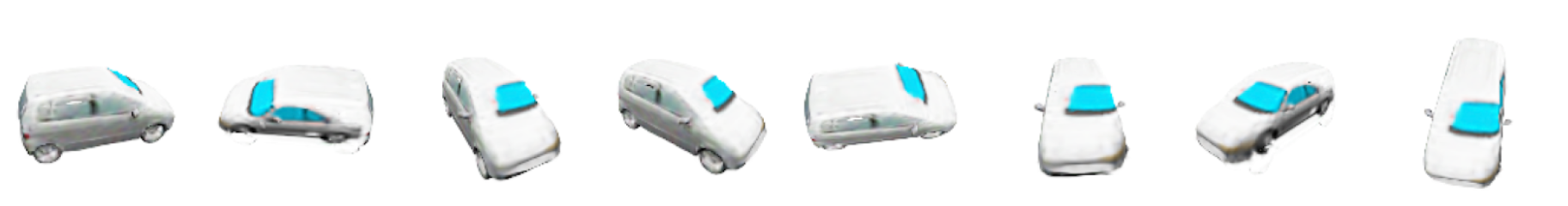}
\adjincludegraphics[width=0.1\linewidth,height=0.1\linewidth,trim={{.125\width} 0 {.75\width} 0},clip]{figures/few_view_recon/fail-test-ours.png}
\adjincludegraphics[width=0.1\linewidth,height=0.1\linewidth,trim={{.25\width} 0 {.625\width} 0},clip]{figures/few_view_recon/fail-test-ours.png}
\adjincludegraphics[width=0.1\linewidth,height=0.1\linewidth,trim={{.375\width} 0 {.5\width} 0},clip]{figures/few_view_recon/fail-test-ours.png}
\adjincludegraphics[width=0.1\linewidth,height=0.1\linewidth,trim={{.5\width} 0 {.375\width} 0},clip]{figures/few_view_recon/fail-test-ours.png}
\adjincludegraphics[width=0.1\linewidth,height=0.1\linewidth,trim={{.625\width} 0 {.25\width} 0},clip]{figures/few_view_recon/fail-test-ours.png}
\adjincludegraphics[width=0.1\linewidth,height=0.1\linewidth,trim={{.75\width} 0 {.125\width} 0},clip]{figures/few_view_recon/fail-test-ours.png}
\adjincludegraphics[width=0.1\linewidth,height=0.1\linewidth,trim={{.875\width} 0 {0\width} 0},clip]{figures/few_view_recon/fail-test-ours.png}
\\
\rotatebox{90}{\quad\ Real} &
\includegraphics[width=0.1\linewidth,height=0.1\linewidth]{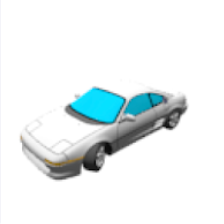} &
\adjincludegraphics[width=0.1\linewidth,height=0.1\linewidth,trim={0 0 {.875\width} 0},clip]{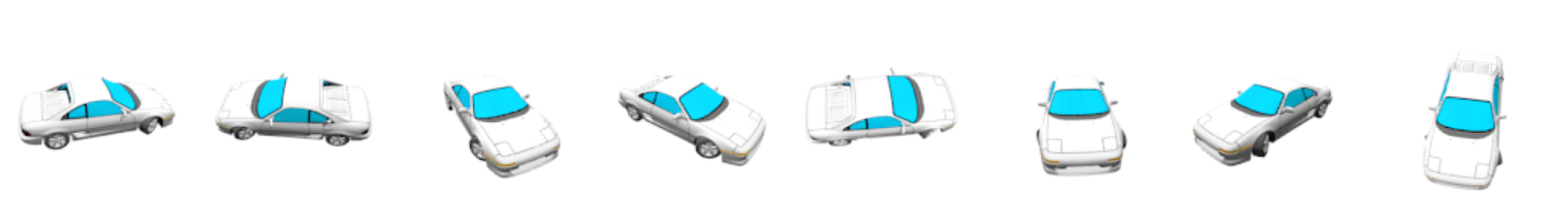}
\adjincludegraphics[width=0.1\linewidth,height=0.1\linewidth,trim={{.125\width} 0 {.75\width} 0},clip]{figures/few_view_recon/fail-test-gtr.png}
\adjincludegraphics[width=0.1\linewidth,height=0.1\linewidth,trim={{.25\width} 0 {.625\width} 0},clip]{figures/few_view_recon/fail-test-gtr.png}
\adjincludegraphics[width=0.1\linewidth,height=0.1\linewidth,trim={{.375\width} 0 {.5\width} 0},clip]{figures/few_view_recon/fail-test-gtr.png}
\adjincludegraphics[width=0.1\linewidth,height=0.1\linewidth,trim={{.5\width} 0 {.375\width} 0},clip]{figures/few_view_recon/fail-test-gtr.png}
\adjincludegraphics[width=0.1\linewidth,height=0.1\linewidth,trim={{.625\width} 0 {.25\width} 0},clip]{figures/few_view_recon/fail-test-gtr.png}
\adjincludegraphics[width=0.1\linewidth,height=0.1\linewidth,trim={{.75\width} 0 {.125\width} 0},clip]{figures/few_view_recon/fail-test-gtr.png}
\adjincludegraphics[width=0.1\linewidth,height=0.1\linewidth,trim={{.875\width} 0 {0\width} 0},clip]{figures/few_view_recon/fail-test-gtr.png} 
\end{tabular}
\vspace{-1.5em}
   \caption{Single-view reconstruction results. The first volume shows the input view. The remaining columns show the output views. The top two rows show a successful example of using our model to create 3D consistent results from single image. The bottom two rows shows typical failure case of the algorithms, where the solution comes from a local optimal where the shape is wrong, but NeRF can still explain the input image well by changing the colors.}
   \label{fig:few_view_recon}
\end{figure*}
We want to show that our diffusion can be used as a prior for a task it has never been supervised on.
In this section, we use single-view reconstruction as the downstream task.
Following the convention from SRN~\cite{SRN}, we sample $251$ images with camera poses in a spiral for each test scene in the dataset and use the 64th image as the input.
We run test-time optimization algorithm~\ref{alg:opt} using equation~\ref{eq:NeRF-color} for $\log p_\theta(y|V)$.
The results are shown in figure~\ref{fig:few_view_recon}, which contains a successful scene and a failure case.
The top row of Figure~\ref{fig:few_view_recon} shows that our model can help obtain a plausible solution that is 3D consistent.
The bottom row shows a typical failure case of our model, which can get stuck in a local minimum where the shape does not match the input image, but the model tries to explain the input image by changing the color.
While hyper-parameter tuning and random reinitialization can alleviate such issues, it remains an interesting challenge for future works.
\section{Conclusion}
Building a generative prior for NeRFs can be useful for a number of downstream tasks.
In this paper, we take the first step to use the diffusion model to learn such a generative prior good for both unconditional and conditional generation tasks.
We show that an important factor in the success of such diffusion prior is making a regularized NeRF dataset.
There are many interesting future work directions to improve the NeRF diffusion generative model.
For example, it will be useful to find a better way to leverage unconditional diffusion prior as guidance for conditional tasks without being stuck with the local minimum of shape mismatch.

\clearpage
\bibliographystyle{iclr2023_conference}
\bibliography{main}

\begin{thebibliography}{38}
\providecommand{\natexlab}[1]{#1}
\providecommand{\url}[1]{\texttt{#1}}
\expandafter\ifx\csname urlstyle\endcsname\relax
  \providecommand{\doi}[1]{doi: #1}\else
  \providecommand{\doi}{doi: \begingroup \urlstyle{rm}\Url}\fi

\bibitem[Bautista et~al.(2022)Bautista, Guo, Abnar, Talbott, Toshev, Chen,
  Dinh, Zhai, Goh, Ulbricht, et~al.]{GAUDI}
Miguel~Angel Bautista, Pengsheng Guo, Samira Abnar, Walter Talbott, Alexander
  Toshev, Zhuoyuan Chen, Laurent Dinh, Shuangfei Zhai, Hanlin Goh, Daniel
  Ulbricht, et~al.
\newblock Gaudi: A neural architect for immersive 3d scene generation.
\newblock \emph{arXiv preprint arXiv:2207.13751}, 2022.

\bibitem[Cai et~al.(2020)Cai, Yang, Averbuch-Elor, Hao, Belongie, Snavely, and
  Hariharan]{ShapeGF}
Ruojin Cai, Guandao Yang, Hadar Averbuch-Elor, Zekun Hao, Serge Belongie, Noah
  Snavely, and Bharath Hariharan.
\newblock Learning gradient fields for shape generation.
\newblock In \emph{Proceedings of the European Conference on Computer Vision
  (ECCV)}, 2020.

\bibitem[Chan et~al.(2021)Chan, Lin, Chan, Nagano, Pan, Mello, Gallo, Guibas,
  Tremblay, Khamis, Karras, and Wetzstein]{EG3D}
Eric~R. Chan, Connor~Z. Lin, Matthew~A. Chan, Koki Nagano, Boxiao Pan,
  Shalini~De Mello, Orazio Gallo, Leonidas Guibas, Jonathan Tremblay, Sameh
  Khamis, Tero Karras, and Gordon Wetzstein.
\newblock Efficient geometry-aware {3D} generative adversarial networks.
\newblock In \emph{arXiv}, 2021.

\bibitem[Chang et~al.(2015)Chang, Funkhouser, Guibas, Hanrahan, Huang, Li,
  Savarese, Savva, Song, Su, Xiao, Yi, and Yu]{ShapeNet}
Angel~X. Chang, Thomas Funkhouser, Leonidas Guibas, Pat Hanrahan, Qixing Huang,
  Zimo Li, Silvio Savarese, Manolis Savva, Shuran Song, Hao Su, Jianxiong Xiao,
  Li~Yi, and Fisher Yu.
\newblock {ShapeNet: An Information-Rich 3D Model Repository}.
\newblock Technical Report arXiv:1512.03012 [cs.GR], Stanford University ---
  Princeton University --- Toyota Technological Institute at Chicago, 2015.

\bibitem[Dhariwal \& Nichol(2021)Dhariwal and Nichol]{diffusionbeatsgan}
Prafulla Dhariwal and Alexander Nichol.
\newblock Diffusion models beat gans on image synthesis.
\newblock \emph{Advances in Neural Information Processing Systems},
  34:\penalty0 8780--8794, 2021.

\bibitem[Dupont et~al.(2022)Dupont, Kim, Eslami, Rezende, and
  Rosenbaum]{functa}
Emilien Dupont, Hyunjik Kim, SM~Eslami, Danilo Rezende, and Dan Rosenbaum.
\newblock From data to functa: Your data point is a function and you should
  treat it like one.
\newblock \emph{arXiv preprint arXiv:2201.12204}, 2022.

\bibitem[Fridovich-Keil et~al.(2022)Fridovich-Keil, Yu, Tancik, Chen, Recht,
  and Kanazawa]{Plenoxels}
Sara Fridovich-Keil, Alex Yu, Matthew Tancik, Qinhong Chen, Benjamin Recht, and
  Angjoo Kanazawa.
\newblock Plenoxels: Radiance fields without neural networks.
\newblock In \emph{Proceedings of the IEEE/CVF Conference on Computer Vision
  and Pattern Recognition}, pp.\  5501--5510, 2022.

\bibitem[Goodfellow et~al.(2020)Goodfellow, Pouget-Abadie, Mirza, Xu,
  Warde-Farley, Ozair, Courville, and Bengio]{GAN}
Ian Goodfellow, Jean Pouget-Abadie, Mehdi Mirza, Bing Xu, David Warde-Farley,
  Sherjil Ozair, Aaron Courville, and Yoshua Bengio.
\newblock Generative adversarial networks.
\newblock \emph{Communications of the ACM}, 63\penalty0 (11):\penalty0
  139--144, 2020.

\bibitem[Graikos et~al.(2022{\natexlab{a}})Graikos, Malkin, Jojic, and
  Samaras]{graikos2022diffusion}
Alexandros Graikos, Nikolay Malkin, Nebojsa Jojic, and Dimitris Samaras.
\newblock Diffusion models as plug-and-play priors.
\newblock \emph{arXiv preprint arXiv:2206.09012}, 2022{\natexlab{a}}.

\bibitem[Graikos et~al.(2022{\natexlab{b}})Graikos, Malkin, Jojic, and
  Samaras]{plug-and-play}
Alexandros Graikos, Nikolay Malkin, Nebojsa Jojic, and Dimitris Samaras.
\newblock Diffusion models as plug-and-play priors.
\newblock \emph{arXiv preprint arXiv:2206.09012}, 2022{\natexlab{b}}.

\bibitem[Ho et~al.(2020)Ho, Jain, and Abbeel]{DDPM}
Jonathan Ho, Ajay Jain, and Pieter Abbeel.
\newblock Denoising diffusion probabilistic models.
\newblock \emph{Advances in Neural Information Processing Systems},
  33:\penalty0 6840--6851, 2020.

\bibitem[Ho et~al.(2022)Ho, Salimans, Gritsenko, Chan, Norouzi, and
  Fleet]{ho2022video}
Jonathan Ho, Tim Salimans, Alexey Gritsenko, William Chan, Mohammad Norouzi,
  and David~J Fleet.
\newblock Video diffusion models.
\newblock \emph{arXiv preprint arXiv:2204.03458}, 2022.

\bibitem[Hoffman et~al.(2022)Hoffman, Le, Sountsov, Suter, Lee, Mansinghka, and
  Saurous]{ProbNeRF}
Matthew~D Hoffman, Tuan~Anh Le, Pavel Sountsov, Christopher Suter, Ben Lee,
  Vikash~K Mansinghka, and Rif~A Saurous.
\newblock Probnerf: Uncertainty-aware inference of 3d shapes from 2d images.
\newblock \emph{arXiv preprint arXiv:2210.17415}, 2022.

\bibitem[Jang \& Agapito(2021)Jang and Agapito]{CodeNeRF}
Wonbong Jang and Lourdes Agapito.
\newblock Codenerf: Disentangled neural radiance fields for object categories.
\newblock In \emph{Proceedings of the IEEE/CVF International Conference on
  Computer Vision}, pp.\  12949--12958, 2021.

\bibitem[Karnewar et~al.(2022)Karnewar, Ritschel, Wang, and Mitra]{ReLUField}
Animesh Karnewar, Tobias Ritschel, Oliver Wang, and Niloy Mitra.
\newblock Relu fields: The little non-linearity that could.
\newblock In \emph{ACM SIGGRAPH 2022 Conference Proceedings}, pp.\  1--9, 2022.

\bibitem[Kingma \& Welling(2013)Kingma and Welling]{VAE}
Diederik~P Kingma and Max Welling.
\newblock Auto-encoding variational bayes.
\newblock \emph{arXiv preprint arXiv:1312.6114}, 2013.

\bibitem[Kosiorek et~al.(2021)Kosiorek, Strathmann, Zoran, Moreno, Schneider,
  Mokr{\'a}, and Rezende]{NeRFVAE}
Adam~R Kosiorek, Heiko Strathmann, Daniel Zoran, Pol Moreno, Rosalia Schneider,
  Sona Mokr{\'a}, and Danilo~Jimenez Rezende.
\newblock Nerf-vae: A geometry aware 3d scene generative model.
\newblock In \emph{International Conference on Machine Learning}, pp.\
  5742--5752. PMLR, 2021.

\bibitem[Kundu et~al.(2022)Kundu, Genova, Yin, Fathi, Pantofaru, Guibas,
  Tagliasacchi, Dellaert, and Funkhouser]{PNF}
Abhijit Kundu, Kyle Genova, Xiaoqi Yin, Alireza Fathi, Caroline Pantofaru,
  Leonidas Guibas, Andrea Tagliasacchi, Frank Dellaert, and Thomas Funkhouser.
\newblock {Panoptic Neural Fields: A Semantic Object-Aware Neural Scene
  Representation}.
\newblock In \emph{CVPR}, 2022.

\bibitem[Mildenhall et~al.(2021)Mildenhall, Srinivasan, Tancik, Barron,
  Ramamoorthi, and Ng]{NeRF}
Ben Mildenhall, Pratul~P Srinivasan, Matthew Tancik, Jonathan~T Barron, Ravi
  Ramamoorthi, and Ren Ng.
\newblock Nerf: Representing scenes as neural radiance fields for view
  synthesis.
\newblock \emph{Communications of the ACM}, 65\penalty0 (1):\penalty0 99--106,
  2021.

\bibitem[M{\"u}ller et~al.(2022{\natexlab{a}})M{\"u}ller, Siddiqui, Porzi,
  Bul{\`o}, Kontschieder, and Nie{\ss}ner]{muller2022diffrf}
Norman M{\"u}ller, Yawar Siddiqui, Lorenzo Porzi, Samuel~Rota Bul{\`o}, Peter
  Kontschieder, and Matthias Nie{\ss}ner.
\newblock Diffrf: Rendering-guided 3d radiance field diffusion.
\newblock \emph{arXiv preprint arXiv:2212.01206}, 2022{\natexlab{a}}.

\bibitem[M{\"u}ller et~al.(2022{\natexlab{b}})M{\"u}ller, Evans, Schied, and
  Keller]{Instant-NGP}
Thomas M{\"u}ller, Alex Evans, Christoph Schied, and Alexander Keller.
\newblock Instant neural graphics primitives with a multiresolution hash
  encoding.
\newblock \emph{arXiv preprint arXiv:2201.05989}, 2022{\natexlab{b}}.

\bibitem[Nichol \& Dhariwal(2021)Nichol and Dhariwal]{nichol2021improved}
Alexander~Quinn Nichol and Prafulla Dhariwal.
\newblock Improved denoising diffusion probabilistic models.
\newblock In \emph{International Conference on Machine Learning}, pp.\
  8162--8171. PMLR, 2021.

\bibitem[Niemeyer et~al.(2022)Niemeyer, Barron, Mildenhall, Sajjadi, Geiger,
  and Radwan]{RegNeRF}
Michael Niemeyer, Jonathan~T Barron, Ben Mildenhall, Mehdi~SM Sajjadi, Andreas
  Geiger, and Noha Radwan.
\newblock Regnerf: Regularizing neural radiance fields for view synthesis from
  sparse inputs.
\newblock In \emph{Proceedings of the IEEE/CVF Conference on Computer Vision
  and Pattern Recognition}, pp.\  5480--5490, 2022.

\bibitem[Poole et~al.(2022)Poole, Jain, Barron, and Mildenhall]{Dreamfusion}
Ben Poole, Ajay Jain, Jonathan~T. Barron, and Ben Mildenhall.
\newblock Dreamfusion: Text-to-3d using 2d diffusion.
\newblock \emph{arXiv}, 2022.

\bibitem[Rombach et~al.(2022)Rombach, Blattmann, Lorenz, Esser, and Ommer]{LDM}
Robin Rombach, Andreas Blattmann, Dominik Lorenz, Patrick Esser, and Bj{\"o}rn
  Ommer.
\newblock High-resolution image synthesis with latent diffusion models.
\newblock In \emph{Proceedings of the IEEE/CVF Conference on Computer Vision
  and Pattern Recognition}, pp.\  10684--10695, 2022.

\bibitem[Saharia et~al.(2022{\natexlab{a}})Saharia, Chan, Chang, Lee, Ho,
  Salimans, Fleet, and Norouzi]{Palatte}
Chitwan Saharia, William Chan, Huiwen Chang, Chris Lee, Jonathan Ho, Tim
  Salimans, David Fleet, and Mohammad Norouzi.
\newblock Palette: Image-to-image diffusion models.
\newblock In \emph{ACM SIGGRAPH 2022 Conference Proceedings}, pp.\  1--10,
  2022{\natexlab{a}}.

\bibitem[Saharia et~al.(2022{\natexlab{b}})Saharia, Ho, Chan, Salimans, Fleet,
  and Norouzi]{S3}
Chitwan Saharia, Jonathan Ho, William Chan, Tim Salimans, David~J Fleet, and
  Mohammad Norouzi.
\newblock Image super-resolution via iterative refinement.
\newblock \emph{IEEE Transactions on Pattern Analysis and Machine
  Intelligence}, 2022{\natexlab{b}}.

\bibitem[Schwarz et~al.(2020)Schwarz, Liao, Niemeyer, and Geiger]{GRAF}
Katja Schwarz, Yiyi Liao, Michael Niemeyer, and Andreas Geiger.
\newblock Graf: Generative radiance fields for 3d-aware image synthesis.
\newblock In \emph{Advances in Neural Information Processing Systems
  (NeurIPS)}, 2020.

\bibitem[Sitzmann et~al.(2019)Sitzmann, Zollh{\"o}fer, and Wetzstein]{SRN}
Vincent Sitzmann, Michael Zollh{\"o}fer, and Gordon Wetzstein.
\newblock Scene representation networks: Continuous 3d-structure-aware neural
  scene representations.
\newblock \emph{Advances in Neural Information Processing Systems}, 32, 2019.

\bibitem[Skorokhodov et~al.(2022)Skorokhodov, Tulyakov, Wang, and
  Wonka]{EpiGRAF}
Ivan Skorokhodov, Sergey Tulyakov, Yiqun Wang, and Peter Wonka.
\newblock Epigraf: Rethinking training of 3d gans.
\newblock \emph{arXiv preprint arXiv:2206.10535}, 2022.

\bibitem[Sohl-Dickstein et~al.(2015)Sohl-Dickstein, Weiss, Maheswaranathan, and
  Ganguli]{sohl2015deep}
Jascha Sohl-Dickstein, Eric Weiss, Niru Maheswaranathan, and Surya Ganguli.
\newblock Deep unsupervised learning using nonequilibrium thermodynamics.
\newblock In \emph{International Conference on Machine Learning}, pp.\
  2256--2265. PMLR, 2015.

\bibitem[Song et~al.(2020)Song, Sohl-Dickstein, Kingma, Kumar, Ermon, and
  Poole]{ScoreSDE}
Yang Song, Jascha Sohl-Dickstein, Diederik~P Kingma, Abhishek Kumar, Stefano
  Ermon, and Ben Poole.
\newblock Score-based generative modeling through stochastic differential
  equations.
\newblock \emph{arXiv preprint arXiv:2011.13456}, 2020.

\bibitem[Sun et~al.(2022)Sun, Sun, and Chen]{DVGO}
Cheng Sun, Min Sun, and Hwann-Tzong Chen.
\newblock Direct voxel grid optimization: Super-fast convergence for radiance
  fields reconstruction.
\newblock In \emph{Proceedings of the IEEE/CVF Conference on Computer Vision
  and Pattern Recognition}, pp.\  5459--5469, 2022.

\bibitem[Tancik et~al.(2021)Tancik, Mildenhall, Wang, Schmidt, Srinivasan,
  Barron, and Ng]{LearnInit}
Matthew Tancik, Ben Mildenhall, Terrance Wang, Divi Schmidt, Pratul~P
  Srinivasan, Jonathan~T Barron, and Ren Ng.
\newblock Learned initializations for optimizing coordinate-based neural
  representations.
\newblock In \emph{Proceedings of the IEEE/CVF Conference on Computer Vision
  and Pattern Recognition}, pp.\  2846--2855, 2021.

\bibitem[Yang et~al.(2019)Yang, Huang, Hao, Liu, Belongie, and
  Hariharan]{PointFlow}
Guandao Yang, Xun Huang, Zekun Hao, Ming-Yu Liu, Serge Belongie, and Bharath
  Hariharan.
\newblock Pointflow: 3d point cloud generation with continuous normalizing
  flows.
\newblock In \emph{Proceedings of the IEEE/CVF international conference on
  computer vision}, pp.\  4541--4550, 2019.

\bibitem[Yu et~al.(2021)Yu, Ye, Tancik, and Kanazawa]{PixelNeRF}
Alex Yu, Vickie Ye, Matthew Tancik, and Angjoo Kanazawa.
\newblock pixelnerf: Neural radiance fields from one or few images.
\newblock In \emph{Proceedings of the IEEE/CVF Conference on Computer Vision
  and Pattern Recognition}, pp.\  4578--4587, 2021.

\bibitem[Zhang et~al.(2023)Zhang, Tang, Niessner, and
  Wonka]{zhang20233dshape2vecset}
Biao Zhang, Jiapeng Tang, Matthias Niessner, and Peter Wonka.
\newblock 3dshape2vecset: A 3d shape representation for neural fields and
  generative diffusion models.
\newblock \emph{arXiv preprint arXiv:2301.11445}, 2023.

\bibitem[Zhou et~al.(2021)Zhou, Du, and Wu]{PointVoxDiff}
Linqi Zhou, Yilun Du, and Jiajun Wu.
\newblock 3d shape generation and completion through point-voxel diffusion.
\newblock In \emph{Proceedings of the IEEE/CVF International Conference on
  Computer Vision}, pp.\  5826--5835, 2021.

\end{thebibliography}


\end{document}